\documentclass[letterpaper, 10 pt, conference]{ieeeconf}  

\IEEEoverridecommandlockouts                              

\usepackage{graphicx}
\usepackage{algorithm}

\usepackage{xcolor}
\usepackage{subcaption}

\usepackage{amsmath}
\usepackage{booktabs}
\usepackage{multirow}
\usepackage[T1]{fontenc}
\DeclareMathOperator*{\argmax}{arg\,max}

\title{\LARGE \bf
Symbolic State Space Optimization for Long Horizon \\Mobile Manipulation Planning}

\author{Xiaohan Zhang$^1$, Yifeng Zhu$^2$, Yan Ding$^1$, Yuqian Jiang$^2$, Yuke Zhu$^2$, Peter Stone$^{2,3}$, Shiqi Zhang$^1$
\thanks{$^1$~Department of Computer Science, The State University of New York at Binghamton \texttt{\{xzhan244; yding25; zhangs\}@binghamton.edu}}
\thanks{$^2$~Department of Computer Science, The University of Texas at Austin \texttt{\{yifeng.zhu@; jiangyuqian@; yukez@cs.; pstone@cs.\}utexas.edu}}
\thanks{$^3$~Sony AI}
}

\begin{document}

\maketitle
\thispagestyle{empty}
\pagestyle{empty}

\begin{abstract}
In existing task and motion planning (TAMP) research, it is a common assumption that experts manually specify the state space for task-level planning.  
A well-developed state space enables the desirable distribution of limited computational resources between task planning and motion planning.  
However, developing such task-level state spaces can be non-trivial in practice.  
In this paper, we consider a long horizon mobile manipulation domain including repeated navigation and manipulation.
We propose Symbolic State Space Optimization~(S3O) for computing a set of abstracted locations and their 2D geometric groundings for generating task-motion plans in such domains.
Our approach has been extensively evaluated in simulation and demonstrated on a real mobile manipulator working on clearing up dining tables. 
Results show the superiority of the proposed method over TAMP baselines in task completion rate and execution time. 
\end{abstract}





\section{Introduction}

At the task level, robots frequently use symbolic planners to sequence high-level actions~\cite{ghallab2016automated}.
At the motion level, each high-level action is grounded to low-level trajectories in continuous spaces using motion planners~\cite{choset2005principles}. 
TAMP algorithms aim to bridge the gap between task planning and motion planning towards enabling robots to fulfill task-level goals and maintain motion-level feasibility at the same time~\cite{lagriffoul2018platform,garrett2021integrated}.
A common and widely accepted assumption for most TAMP research is that the task planner is predefined by a domain expert who manually specifies a symbolic state space.
In this paper, we discuss TAMP in a long horizon mobile manipulation domain where the robot is given a task of repeated navigation to perform manipulation behaviors (e.g., pick and place) in different places. 
 

    
\begin{figure}
\begin{center}
    \includegraphics[width=0.5\textwidth]{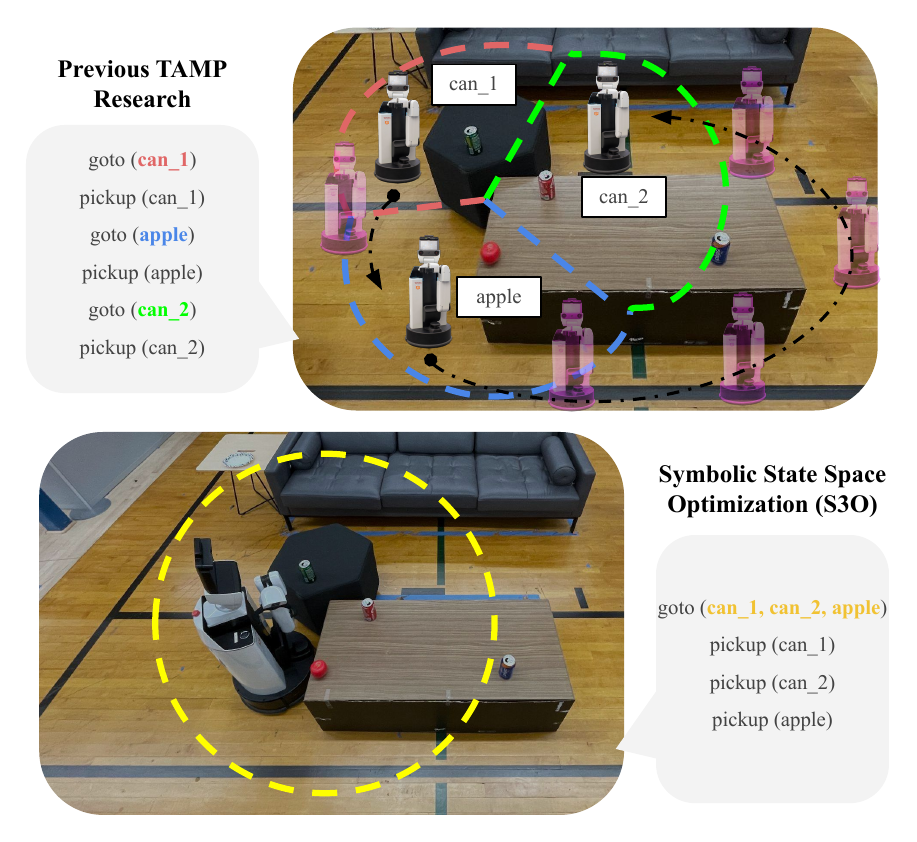}
    \caption{Objects are frequently in separate symbolic locations in a predefined task planner. A TAMP system with such a fine-grained state space would always generate plans that suggest the robot navigate before every manipulation. However, if an optimized state space can include multiple objects (that are close to each other) in a single location, the robot will be able to navigate once and perform a sequence of manipulation actions. We aim to answer how to compute such symbolic locations and their geometric groundings. 
    }
    \label{fig:domain}
\end{center}
\end{figure}

Nevertheless, manually constructing state spaces might not be desirable in some scenarios.
Fig.~\ref{fig:domain} shows a situation where in long horizon mobile manipulation domains, if each object is placed at a separate symbolic location as defined in the task-level state space, the robot will always need to navigate before picking up the next object.
This is because the task planner believes only a navigation action can bring the robot to the location required by the next manipulation action. 
In practice, however, the robot often picks up multiple objects from a single position, for example, as restaurant waiters can easily identify a standing location that allows them to pick up multiple dishes at once.  
Especially when objects are located close to each other, it is unnecessary for the robot to navigate before every manipulation. 
This observation motivated the development of this research on optimizing symbolic state spaces for task planners to best facilitate TAMP for long horizon mobile manipulation. 

One of the challenges in optimizing symbolic state spaces for TAMP, which is the focus of this work, comes from the uncertainties in action and perception. 
We consider failures in navigation and manipulation behaviors, e.g., due to the robot being too close to obstacles or too far from the target objects.
To this end, we propose \textbf{S}ymbolic \textbf{S}tate \textbf{S}pace \textbf{O}ptimization~(S3O) based on probabilistically evaluated action feasibility under uncertainty.
S3O partitions the continuous configuration space into a set of abstracted locations with their 2D geometric groundings to compute efficient and feasible task-motion plans in long horizon mobile manipulation domains.

\begin{figure}
\begin{center}
    \includegraphics[width=0.48\textwidth]{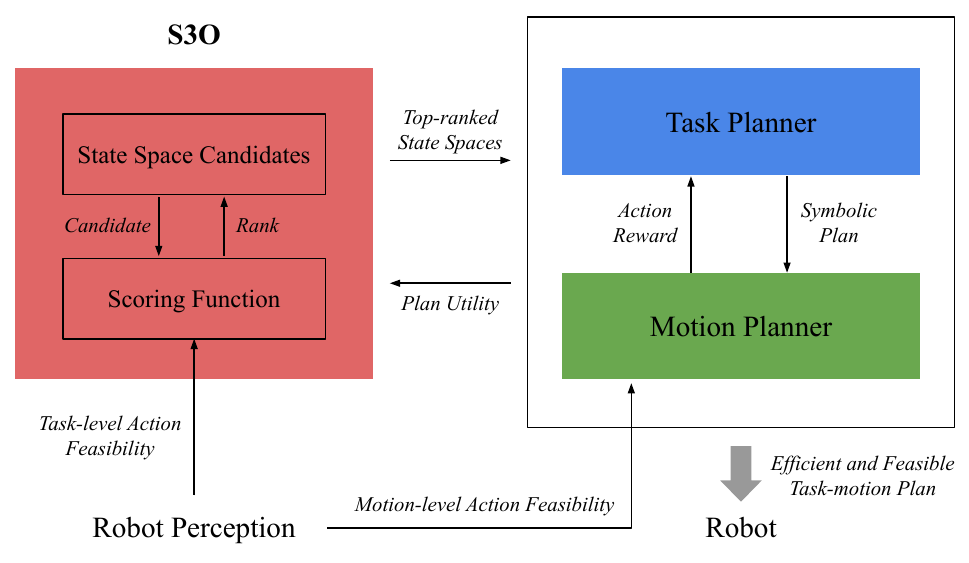}
    \caption{An overview of Symbolic State Space Optimization~(S3O) for Task and Motion Planning systems. 
    }
    \label{fig:s3o_overview}
\end{center}
\end{figure}

Fig.~\ref{fig:s3o_overview} shows an overview of S3O which first constructs a candidate set of object-centric symbolic state spaces using Voronoi Partitioning~\cite{okabe1997locational}.
Then the algorithm ranks each state space by a scoring function developed using feasibility evaluation from robot perception.
The ranking mechanism effectively reduces the search complexity of state spaces by controlling the size of the candidate set.
Top-ranked state spaces are used for constructing the task planner in the TAMP system where we further apply an Evolution Strategy~(ES) algorithm~\cite{hansen2003reducing} for efficient motion-level search.
The proposed framework has been quantitatively evaluated in simulation and qualitatively demonstrated on real hardware, where the robot works on the task of ``clearing up dining tables.''
Compared to existing TAMP baselines, experimental results show that our approach consistently leads to high-quality task-motion plans in terms of task completion rate and plan execution time.


\section{Related Work}
In this section, we describe the three most related research areas, including those task and motion planning (TAMP) methods that optimize plan efficiency and feasibility, research that learns symbolic representations for robot planning, and the application domain of long horizon mobile manipulation. 

\subsection{TAMP for Efficient and Feasible Behaviors}
Task and motion planning research can be categorized into two groups: one includes high-level actions that take no more than a few seconds (e.g., picking up, putting down and pushing objects), the other requires robot actions taking relatively long time (e.g., long-term navigation)~\cite{lo2020petlon}.
The former type of TAMP has a long history in the literature and focuses mostly on action feasibility~\cite{toussaint2015logic, zhu2020hierarchical, garrett2018ffrob, wells2019learning, migimatsu2020object, mcmahon2017robot, zhao2018reactive}, while some recent methods have considered behavioral efficiency in the latter type of TAMP, usually in robot navigation, autonomous driving, or mobile manipulation domains~\cite{lo2020petlon,jiang2019task,ding2020task,thomas2021mptp,zhang2022visually, ding2022glad, ding2023task}.
One common assumption for these TAMP methods is the predefined task planner.
Unlike those methods, we probabilistically compute action feasibility via visual perception to optimize the state space of the task planner. 


\subsection{Symbol Learning for Robot Planning}
Learning-based methods have shown effectiveness in model acquisition and symbol generation for robot planning.
Researchers have learned action preconditions and effects models for enabling purely symbolic planning~\cite{konidaris2018skills} and integrated task-motion planning~\cite{wang2021learning}.
Some other work focuses on symbol learning and mapping, such as connecting natural language to learned symbolic abstractions~\cite{gopalan2020simultaneously}, learning state abstractions for bootstrapping motion planning~\cite{shah2022using}, and learning to ground the physical meanings of object attribute symbols in the real world~\cite{ding2022learning}.
In our work, we also learn to generate and map each symbol from the continuous space, but going beyond that, we further optimize the efficiency and feasibility of task-motion plans for the robot to execute using the learned symbols.

\subsection{Long Horizon Mobile Manipulation}
There is rich literature on learning and planning coordinated actions for mobile manipulation~\cite{yamamoto1992coordinating, thakar2023survey}. 
Most existing methods focused on positioning the base of a mobile manipulator in such a way that manipulability is maximized~\cite{berenson2008optimization,diankov2008openrave,stulp2012learning,ren2016method}.
A convincing technique is using robot reachability maps~\cite{zacharias2008positioning,vahrenkamp2013robot}. 
Recent research applies reinforcement learning in a hierarchical style to tackle this problem~\cite{jauhri2022robot, gu2022multi,xia2020relmogen,li2020hrl4in, lew2022robotic}.
In this paper, we not only consider coordinated navigation and manipulation, but also optimize a sequence of mobile manipulation actions over a long horizon. Sequential mobile manipulation tasks~\cite{carriker1991path} have been studied, including works that aimed to minimize platform movements to reach a set of poses in the workspace~\cite{xu2020planning} or to minimize the overall cost of completing the task~\cite{reister2022combining}. 
As compared to our approach, we also consider perception and assume execution-time uncertainty from both perception and actuation.

\section{Problem Statement}
\label{sec:problem}
We present the terminologies, assumptions, and objectives of the TAMP problem we focus on in this research: a long horizon mobile manipulation task where the robot repeatedly navigates and picks up multiple objects in different locations.
\vspace{0.5em}

\noindent\textbf{Symbols and Symbol Mapping: }
$\mathcal{O} = \{o_1,o_2 ...\}$ is a set of target objects that can be moved by a robot.
$\mathcal{L} = \{l_1,l_2 ... \}$ is a set of symbolic locations.
Let $y \in \mathcal{Y}$ be a set of xy poses in continuous space.
$Sym: \mathcal{Y}\rightarrow \mathcal{L}$ is a function that maps any 2D geometric position $y \in \mathcal{Y}$ to a symbolic location $l \in \mathcal{L}$.
The symbolic state space of our problem is defined in the form of $\langle \mathcal{L}, Sym, \mathcal{Y} \rangle$. 


\noindent\textbf{Actions: }
The robot is equipped with skills of performing a set of actions denoted as $\mathcal{A}: \mathcal{A}^n \cup \mathcal{A}^m$, where $\mathcal{A}^n$ and $\mathcal{A}^m$ are \emph{navigation} actions and \emph{manipulation} actions respectively.
A navigation action 
$a^n: \langle l_r, l'_r, y_r, y'_r \rangle \in \mathcal{A}^n$ 
is specified at both low and high levels: 1) the robot's current and next symbolic locations that are denoted as $l_r, l'_r\in \mathcal{L}$; 2) the corresponding 2D coordinates $y_r, y'_r$ mapped by $Sym$.
$r$ is a symbol to denote ``robot'' as being distinguished from symbol $o$ for ``object''.
A manipulation action
$a^m: \langle o, l, y_r, y_o \rangle \in \mathcal{A}^m$
is specified by an object (i.e., $o$) to be manipulated, the object's 2D location, $y_o$, the object's and the robot's symbolic location, $l \in \mathcal{L}$.
The robot and the object to be manipulated should be in the same symbolic location.
In this work, we consider \texttt{pickup} as a manipulation action and \texttt{goto} as a navigation action.
Actions are defined via preconditions and effects. 
For instance, the action \texttt{pickup($o_1$)} has preconditions of \texttt{at(robot, $l_1$)} and \texttt{at($o_1$, $l_1$)}, meaning that to pick up the object $o_1$, the object must be co-located with the robot base in the same symbolic location $l_1$.
The effects of \texttt{pickup($o_1$)} include $o_1$ being moved into the robot’s hand, i.e., \texttt{inhand($o_1$)}.

\vspace{0.5em}
\noindent\textbf{Action Uncertainty: }
Let $\mathcal{T}: \mathcal{T}^n \cup \mathcal{T}^m$ be a set of probability distributions for modeling action uncertainties.
For a navigation action,
$\mathcal{T}^n(\hat{y}'_r|y_r, y'_r)$ represents the probability of a mobile robot aiming to navigate to goal $y'_r$, while landing in $\hat{y}'_r$, given the current robot position $y_r$.
For a manipulation action, $\mathcal{T}^m(\hat{y}'_o|y_o,y_r)$
represents the probability of the robot given an end effector goal position $y_o$ of ``reaching'' the object, while ending up at a position $\hat{y}'_o$, given the robot's standing position $y_r$.
In practice, $\mathcal{T}^n$ and $\mathcal{T}^m$ are determined by the robot’s navigation and manipulation systems.
In this paper, both $\mathcal{T}^n$and $\mathcal{T}^m$ are treated as black box.

\vspace{0.5em}
\noindent\textbf{Perception: }
The robot visually perceives the environment. 
While we provide the robot with top-down view images in this work, our approach can be combined with perception methods that rely on first-person view for object pose estimation~\cite{wang2019densefusion}. 
A map is generated in a pre-processing step, and provided to the robot as prior information for navigation purposes using rangefinder sensors. 
Please note that dynamic obstacles such as randomly-placed chairs are not in the map. 

\vspace{0.5em}
\noindent\textbf{Problem Formulation: }
The input of the problem is a tuple $\langle \mathcal{Y}_o^{init}, y_r^{init}, \mathcal{A} \rangle$.
$\mathcal{Y}_o^{init}$ is a set of objects' initial positions and $y_r^{init}$ is the robot's initial base pose.
The problem outputs a task-motion plan $p$ which is in the form of a sequence of navigation actions $a^n \in A^n$ and manipulation actions $a^m \in A^m$.
The problem finds a task planner $Pln^{\mathcal{L}, Sym,\mathcal{Y}}$ that is parameterized by the symbolic state space $\langle L, Sym, \mathcal{Y}\rangle$, in order to compute a task-motion plan $p$, where the objective is to maximize the plan utility for improving task completion rate and reducing robot execution time. 

\begin{figure}
\begin{center}
    \includegraphics[width=0.49\textwidth]{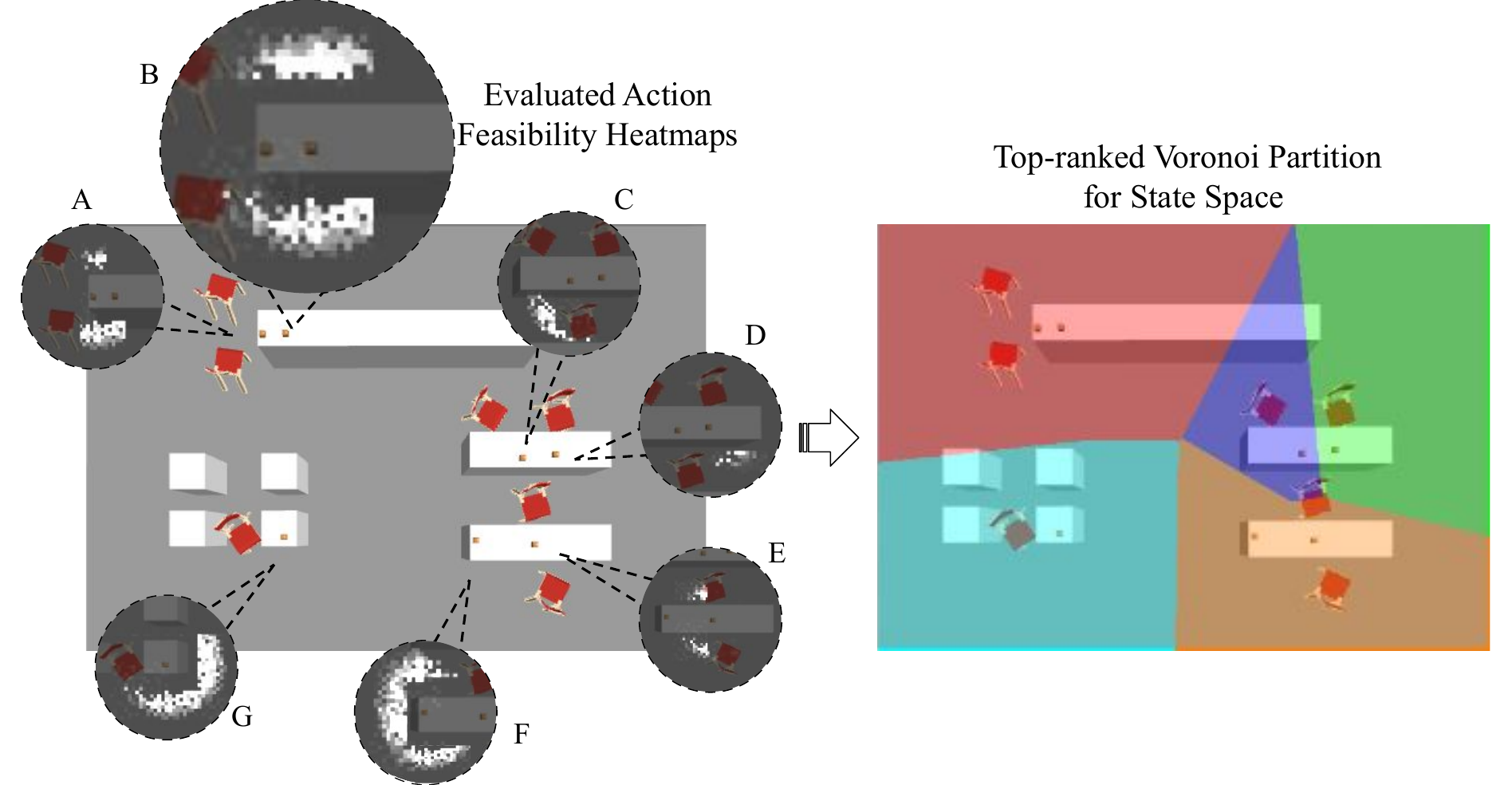}
    \caption{\textbf{Left:} Action feasibility values are computed using robot perception and represented as heatmaps. \textbf{Right:} A top-ranked Voronoi Partition for the state space generated using S3O, where objects A and B are in one symbolic location, and objects E and F are in another one.
    }
    \label{fig:feasibility}
\end{center}
\end{figure}
\section{Symbolic State Space Optimization (S3O)}
In this section, we present the paper's main contribution called Symbolic State Space Optimization (S3O) which optimizes the state space for the task planner based on probabilistically evaluated action feasibility.
S3O first constructs symbolic state spaces using object-centric Voronoi Partitioning and robot reachability.
And then it ranks a set of candidate state spaces based on evaluated action feasibility under uncertainty. 


\vspace{0.5em}
\noindent\textbf{Constructing Symbolic State Spaces: }
We construct state spaces following two principles: 1) states (i.e., locations) should be determined by which object(s) they are the closest to; 2) the distance from the object to each pose in a state should be within the maximum reachability (1 meter in our case) of the robot.
 Thus, in our framework, we consider poses that are around the objects within 1 meter, and generate areas by object positions in the 2D configuration space using the Voronoi Partitioning algorithm.
The distance from each 2D pose in an area to its corresponding object position is less than that from every other object position. 
Each area in the Voronoi diagram is considered as a symbolic location $l$, and the whole Voronoi partition corresponds to a set of locations $\mathcal{L}$ as well as a symbol mapping function $Sym$ to map each 2D pose to a location $l \in \mathcal{L}$.
Further, possible adjacency area merging operations are conducted in the Voronoi diagram.
Each area merging operation that results in a new symbolic state space (i.e., $\langle \mathcal{L}, Sym, \mathcal{Y}\rangle$) is considered as a state space candidate.


\vspace{0.5em}
\noindent\textbf{Scoring Function for State Space Ranking: }
In order to deal with a large number of objects, we compute scores for each state space candidate, i.e., $\langle \mathcal{L}, Sym, \mathcal{Y}\rangle$.
The score is calculated using the following function that is based on action feasibility:
\begin{align}
    \textit{Score}(\langle \mathcal{L}, Sym, \mathcal{Y} \rangle) = \sum_{o \in \mathcal{O}}{Fea}^t(l, o), \textnormal{if \texttt{at($o$, $l$)}}
    \label{eqn:score}
\end{align}
where ${Fea}^t(l, o)$ is the task-level action feasibility function that computes the probability of the robot navigating to location $l$ and picking up object $o$. 
Intuitively, if the symbolic state space has a high accumulative task-level feasibility value over all the objects, this state space will be evaluated with a high score.
Fig.~\ref{fig:feasibility} shows the evaluated action feasibility (represented as heatmaps) and a top-ranked Voronoi Partition for the state space. 

After ranking the state spaces by the scores computed using Eqn.~\ref{eqn:score}, we select the top $K$ state spaces to construct $K$ task planners at robot planning time.
In each TAMP search iteration, our system normalizes the scores to produce a probability distribution from which one of the task planners is chosen.
The system plans in parallel, each with a sampled task planner, and uses $\argmax$ to find the state space (i.e., $\langle \mathcal{L}, Sym, \mathcal{Y}\rangle$) that generates a plan of the highest utility.

\vspace{0.5em}
\noindent\textbf{Action Feasibility Evaluation: } Robot perception is used to probabilistically evaluate action feasibility, represented as function $Fea: {Fea}^t \cup {Fea}^m$.
The task-level feasibility function ${Fea}^t(l, o)$ takes a symbolic location $l$ and an object $o$ as input, while the motion-level feasibility function ${Fea}^m(y_r, y_o)$ takes a robot 2D pose $y^r$ and an object 2D pose $y^o$ as input. Both task-level and motion-level feasibility functions output feasibility values ranging from 0.0 (infeasible) to 1.0 (feasible).
In this work, ${Fea}^t$ serves as a key component in the proposed scoring function (Eqn.~\ref{eqn:score}).
${Fea}^m$ is used to compute: 1) ${Fea}^t$, which is discussed in the next paragraph, and 2) the plan utility, which is formally defined in the next section.

Our task-level feasibility function ${Fea}^t(l, o)$ shares the same definition as what was initially introduced in~\cite{zhang2022visually}. 
Briefly summarizing here, ${Fea}^t(l, o)$ relies on ${Fea}^m(y_r, y_o)$ and a sampling function $Smp$.
${Fea}^m(y_r, y_o)$ computes the motion-level feasibility of robot navigating to 2D position $y_r$ and picking up the object that is at position $y_o$. 
$Smp$ samples 2D positions $y_r$ that satisfy $Sym(y_r) = l$, where the positions are weighted by ${Fea}^m(y_r, y_o)$.
In other words, positions of higher motion-level feasibility are more likely to be sampled. 
Computing ${Fea}^t(l, o)$ is to calculate the average motion-level feasibility over $N$ samples that are drawn using $Smp$.

We extract ${Fea}^m(y_r, y_o)$ from a learned Fully Convolutional Network model~\cite{long2015fully}, which is trained using robot data from past experience, represented as gray-scale heatmap images.
We trained the model by collecting a dataset that diversifies the obstacle (i.e., chair) positions and is with randomly-placed objects on the table.
One recent work uses the same architecture for motion-level feasibility evaluation~\cite{zhang2022visually}, but their model can only deal with objects that are of a predefined distance from the table edge due to the limitation of its training dataset.
In comparison, the motion-level feasibility function extracted from our model equips the robot with the capability of handling more generalized object pick and place tasks.



\begin{figure}
     \centering
     \begin{subfigure}{0.23\textwidth}
         \centering
         \includegraphics[width=\textwidth]{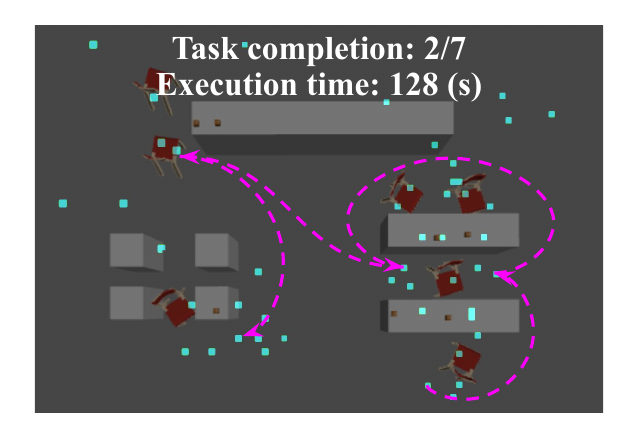}
         \caption{Early iteration.}
     \end{subfigure}
     \begin{subfigure}{0.23\textwidth}
         \centering
         \includegraphics[width=\textwidth]{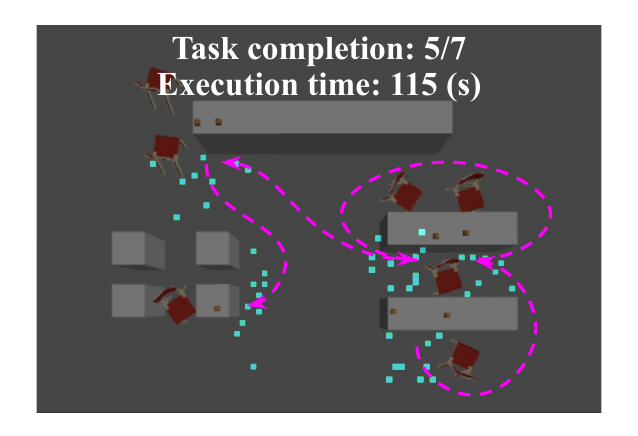}
         \caption{Late iteration.}
     \end{subfigure}

     \caption{Samples (cyan pixels) drawn from the CMA-ES sampler at early and late iterations. With action feasibility and efficiency being considered in the objective function, robot base positions gradually converge to a sequence of areas that are close to the objects for the robot to reach, and are of a low overall navigation cost. 
     }
    \label{fig:cma-es}
\end{figure}

\section{Computing Task-motion Plans}
\label{sec:motion}
Long horizon mobile manipulation domains require robots to complete tasks as accurately and quickly as possible.
This subsection details how we use the optimized state spaces from S3O to compute feasible and efficient task-motion plans.

As described in Sec.~\ref{sec:problem}, the objective of the problem is to maximize the overall task completion rate and minimize the robot execution time.
Robot execution time is largely affected by how much time each action takes (especially long-range navigation actions), and the task completion rate depends on manipulation action feasibility. 
To this end, at planning time, we design the cost function for an action $a$ as:
\begin{equation}
    Cst(a)=\begin{cases}
      len(y_r, y'_r) / v + \gamma, \textnormal{ if $a\in \mathcal{A}^n$}\\
      \delta, \textnormal{ if $a\in \mathcal{A}^m$}
    \end{cases}    
\end{equation}
where function $len$ is able to measure the trajectory length of executing a navigation action and $v$ is the robot speed.
$\gamma$ is a constant cost for navigation when the robot starts moving, which motivates the robot to select as few navigation actions as possible.
$\delta$ is a constant cost for manipulation actions which is relatively small as compared to the cost for navigation actions.

We further use the action cost function to design the action reward function. 
Let $\lambda$ be a successful reward bonus of picking up one object.
The action reward function is designed as follows:
\begin{equation}
R(a)=\begin{cases}
      -Cst(a), \textnormal{if $a\in \mathcal{A}^n$}\\
      -Cst(a)+Fea^m(y^r, y^o) \cdot \lambda, \textnormal{if $a\in \mathcal{A}^m$}
    \end{cases}
\end{equation}

We use the CMA-ES optimization technique~\cite{hansen2003reducing} to serve as the sampling algorithm for motion-level 2D poses that the robot navigates to and performs the manipulation action(s) at.
Fig.~\ref{fig:cma-es} shows an example of the samples drawn from early and late iterations of the CMA-ES sampler. 
Each sample we draw is in the form of $\langle y_r^1[x], y_r^1[y], y^2_r[x], y_r^2[y],... \rangle$, where $y_r^{i}[x]$ ($y_r^{i}[y]$) denotes the $x$ ($y$) coordinate of the robot pose for navigating to and picking up the $i$th object from.
We maintain an independent CMA-ES sampler for each fixed task-level sequence, so we are able to form a complete task-motion plan $p$ by simply chaining the sampled $xy$ positions.
Two consecutive pairs of $xy$ positions (i.e., $\langle y_r^{i}[x],y_r^{i}[y] \rangle$ and $\langle y_r^{i+1}[x],y_r^{i+1}[y] \rangle$) can be used to parameterize a navigation action, and every single pair of $xy$ positions plus an object position (i.e., $y_o$) can be used to parameterize a manipulation action.
This enables us to convert a sample to a sequence of actions and then evaluate the sample by computing $\sum R(a)$.
$\sum R(a)$ is the utility of a task-motion plan and serves as the objective function for the CMA-ES sampler.

\begin{figure*}
     \centering
     \begin{subfigure}{0.47\textwidth}
         \centering
         \includegraphics[width=\textwidth]{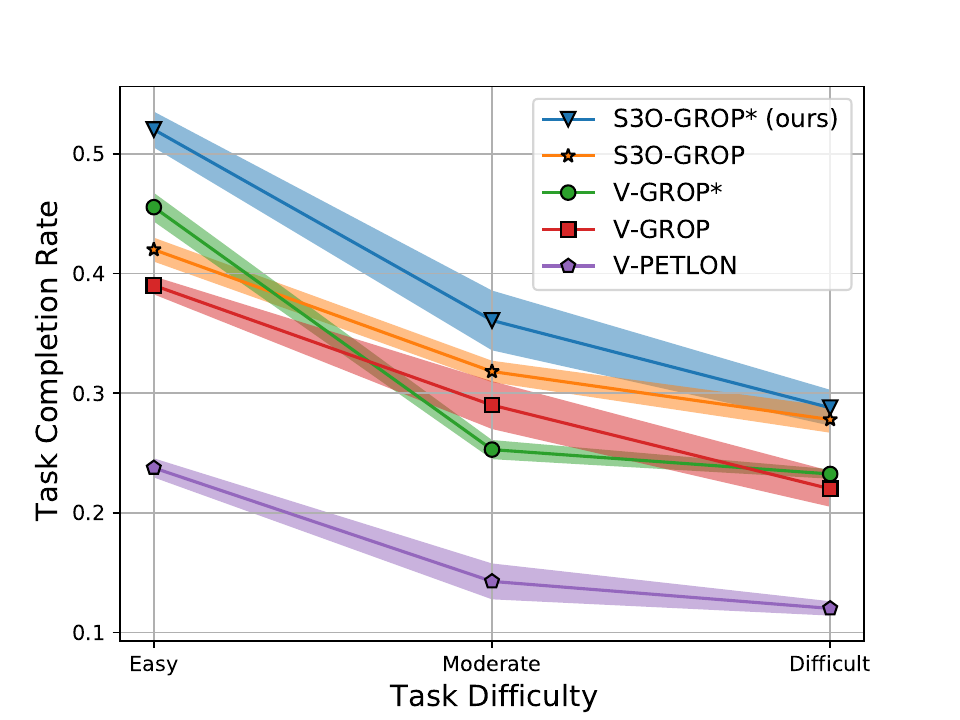}
         \caption{Task completion rate.}
     \end{subfigure}
     \begin{subfigure}{0.47\textwidth}
         \centering
         \includegraphics[width=\textwidth]{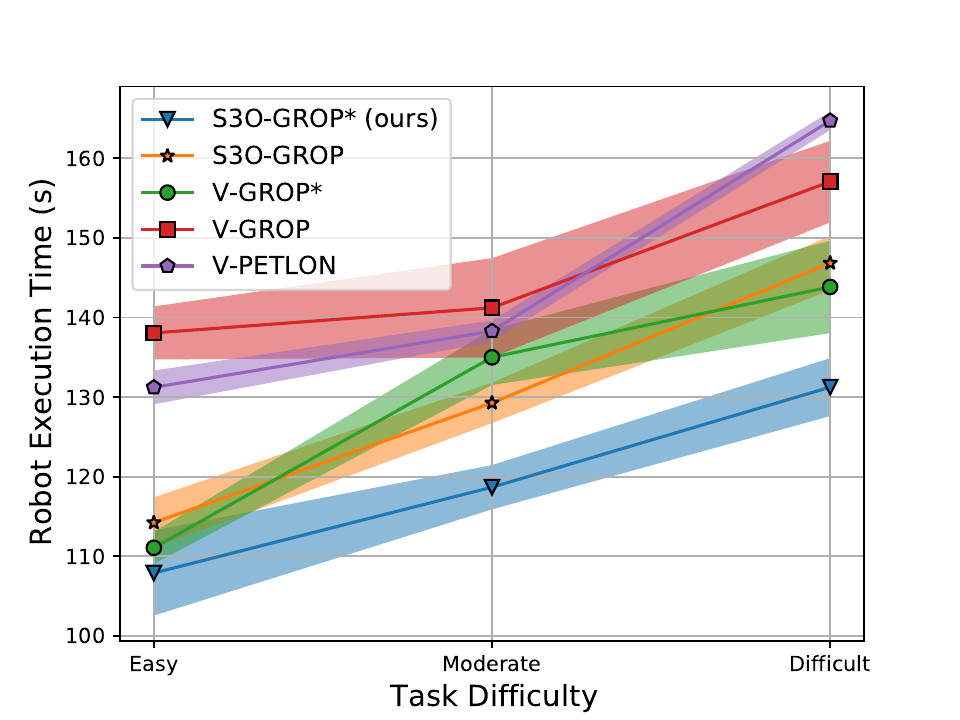}
         \caption{Robot execution time.}
     \end{subfigure}

     \caption{Overall performances of our approach (S3O-GROP$^*$) and four baseline methods in task completion rate and robot execution time (s). 
    Tasks are grouped based on their difficulties.
    S3O-GROP$^*$ produced the highest task completion rate while maintaining the lowest robot execution time. 
    This observation is consistent over tasks of different difficulties. 
    }
    \label{fig:exp_main}
\end{figure*}

\section{Experiments}
We conducted extensive experiments in simulation, where a mobile manipulator performs navigation and manipulation actions to ``collect dishes'' in a ``restaurant'' scenario. 
We also demonstrated the computed plan using our method on a real robot system. 
Our main hypothesis is that under a planning time budget, the proposed framework outperforms existing TAMP algorithms in task completion rate and robot execution time. 




\subsection{Baselines}
Ours and the baseline methods differ from each other in how to construct and optimize task planners (state spaces in particular).
We compare S3O with basic object-centric Voronoi Partitioning (denoted as ``V'').
After selecting a task planner, there are different TAMP strategies we can choose from.
We consider two TAMP algorithms for navigation domains from the literature, which are GROP~\cite{zhang2022visually} and PETLON~\cite{lo2020petlon}.
Our TAMP component is built on GROP and further incorporates the proposed CMA-ES sampling algorithm for motion-level optimization.
Thus, we denote our TAMP strategy as GROP$^*$.
Combining different methods from task planner construction and TAMP strategy, we consider the following five methods in total: S3O-GROP$^*$, S3O-GROP, V-GROP$^*$, V-GROP, and V-PETLON.
To improve clarity, we briefly summarize the major differences between the five methods:
\begin{itemize}
    \item \textbf{S3O-GROP$^*$} (proposed): It optimizes state spaces using S3O, and samples navigation goals using CMA-ES. The algorithm optimizes both plan efficiency and action feasibility.
    \item \textbf{S3O-GROP}: One ablative version of S3O-GROP$^*$.
    It is the same as S3O-GROP$^*$ except that CMA-ES is not used here.
    \item \textbf{V-GROP$^*$}: One ablative version of S3O-GROP$^*$. 
    It is the same as S3O-GROP$^*$ except that there is no state space optimization.
    \item \textbf{V-GROP}~\cite{zhang2022visually}: It does not optimize the state space, and samples navigation goals only by evaluated feasibility. The algorithm optimizes both plan efficiency and action feasibility.
    \item \textbf{V-PETLON}~\cite{lo2020petlon}: It does not optimize the state space, and selects navigation goals by just randomly sampling an obstacle-free position that is close to the object position. The algorithm optimizes plan efficiency but does not evaluate action feasibility.
\end{itemize}
In comparison, our method, S3O-GROP$^*$, constructs the task planner using S3O and selects navigation goals by CMA-ES sampling, whose objective includes both motion-level feasibility and long horizon mobile manipulation cost. 
Note that we did not include ``S3O-PETLON'' as one of the baselines as there is no feasibility evaluation in the original PETLON algorithm, thus S3O is inapplicable.

\subsection{Experimental Setup}
The simulation environment contains seven tables of different sizes: one long table as the ``bar area'', two mid-sized tables, and four small tables that are able to take one person per table.
Objects to be collected are randomly generated on the tables, and an obstacle (i.e., chair) that is not mapped beforehand is placed near each object with a randomly generated position and orientation.
The number of objects is dynamically changed for different environments, ranging from 5 to 7.
An RGB camera is attached to the ceiling to capture overhead images of environments for robot perception. 
We assume the robot can hold multiple objects at the same time. 
Task completion is evaluated based on whether ``dishes'' on the tables are successfully ``collected'' or not. 

The mobile manipulator in simulation includes a UR5e robot arm, a Robotiq 2F-140 gripper, an RMP 110 mobile base, and a Velodyne VLP-16 lidar sensor on the mobile base. 
We used the Building-Wide Intelligence (BWI) codebase~\cite{khandelwal2017bwibots} to construct our simulation platform, which relies on the Gazebo physics engine~\cite{koenig2004design}.
Rapidly exploring Random Tree (RRT) approach~\cite{lavalle1998rapidly} is used to compute motion-level manipulation plans. 
The navigation stack was built using the \texttt{move\_base} package of Robot Operating System (ROS)~\cite{quigley2009ros}.
The robot's task planner is ASP-based~\cite{gelfond2014knowledge,lifschitz2002answer} and the Clingo solver is applied for computing task plans~\cite{gebser2014clingo}.
 We adopted the FCN-VGG16 model~\cite{long2015fully} for predicting action feasibility heatmaps.
The model is trained using a machine equipped with an Intel 3.80GHz i7-10700k CPU and a GeForce RTX 3070 GPU on a Ubuntu system.

\subsection{Planning Parameters}
At planning time, we do parallel computing using 12 CPUs on a different machine from training the FCN model.
The machine for planning is equipped with an 11th Gen Intel(R) 2.30GHz Core(TM) i7-11800H CPU.
The planning time budget is set to 300 seconds.
For each task-level sequence, the maximum number of motion-level samples that can be drawn is 200.
The manipulation constant cost $\delta$ is set to 5, and the navigation constant starting cost $\gamma$ is set to 20.
The reward for a successful manipulation action $\lambda$ has a value of 150.
The robot velocity $v$ for computing action cost is set to 0.4m/s.
There are other parameters for the CMA-ES sampler. 
We consider the first 20 generations for CMA-ES sampling.
Since the number of motion-level samples is fixed (i.e., 200), the population size of each generation is set to 10.
After ranking all possible state spaces, we choose the top 5 of them according to the computed scores.


\subsection{Task Completion Rate and Robot Execution Time}
Fig.~\ref{fig:exp_main} shows the main results of task completion rate and robot execution time.
There were a total of 100 different tasks.
We grouped the tasks based on their difficulties: Easy, Moderate, and Difficult. 
A task's \emph{difficulty} is measured by the total area that a robot can navigate to and pick up an object from. 
For instance, a task with all feasible picking up positions being surrounded by obstacles has a high difficulty. 
After sorting the tasks based on their difficulties, we evenly placed them into the three groups. 

Our system consistently performed the best in task completion rate (left subfigure) in all three settings, while maintaining the lowest robot execution time (right subfigure). 
We also see that methods that use S3O (i.e., S3O-GROP$^*$ and S3O-GROP) have better or at least similar performance compared with the methods that use basic Voronoi Partitioning (i.e., V-GROP$^*$, V-GROP, and V-PETLON).
While only considering methods that use Voronoi Partitioning, the one that uses GROP$^*$ generates plans that are of the least execution time and maintains a similar (higher) success rate as compared to V-GROP (V-PETLON).
Both GROP$^*$ and GROP consider feasibility when sampling navigation points, but the former also takes efficiency into account by using CMA-ES.
That is the reason why V-GROP$^*$ and V-GROP share similar success rates but the former performs better in plan efficiency.
Overall, the results support our hypothesis.

\begin{table}

\centering
\caption{Ablation study on the impact of different strategies for constructing the task planner. \textbf{Task completion rate / robot execution time} are reported in the table. S3O is our method that does task planner optimization; S3O-Random is an ablative version that uniformly selects the task planner from the candidate set without score ranking.}
\scriptsize

\begin{tabular}[t]{@{}lcc@{}}
	\toprule
	 Task Difficulty & S3O & S3O-Random\\ \midrule
	 Easy & \textbf{0.52} $\pm$ 0.02 / 107.90 $\pm$ 5.35 & 0.30 $\pm$ 0.06 / \textbf{96.95} $\pm$ 2.40\\ \midrule
      Moderate & \textbf{0.36} $\pm$ 0.03 / 118.68 $\pm$ 2.78 & 0.18 $\pm$ 0.08 / \textbf{97.89} $\pm$ 4.22\\ \midrule
	 Difficult & \textbf{0.29} $\pm$ 0.02 / 131.25 $\pm$ 3.64 & 0.17 $\pm$ 0.08 / \textbf{102.85} $\pm$ 6.30\\ 
\bottomrule
\end{tabular}
\label{tab:ablation}
\end{table}

\begin{figure*}
\begin{center}
    \includegraphics[width=\textwidth]{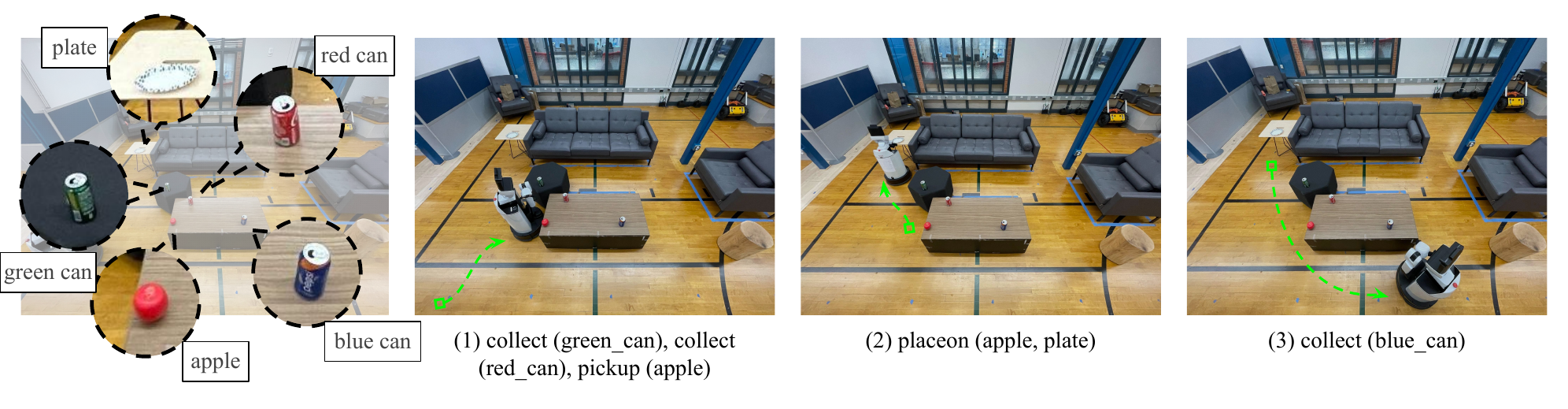}
    \caption{Real robot demonstration of the planned trajectory computed using our optimized task planner. 
    }
    \label{fig:real}
\end{center}
\end{figure*}

\subsection{Ablation Study}
We also conducted an ablation study (as shown in TABLE~\ref{tab:ablation}) to learn the impact of different strategies for constructing the task planner, specifically how to select a state space from a set of state space candidates at planning time.
Without a predefined state space, we compare two methods for state space selection: the proposed S3O (with score ranking), and an approach that uniformly samples state spaces from all possible candidates (denoted as ``Random'').
We observe that by considering S3O, the robot achieves a higher task completion rate for all tasks.
When uniformly selects a state space to construct the task planner, the system produces more cost-efficient plans, however, suffers from very poor performance in completing the task.
The reason is that the random selection strategy treats every state space candidate equally, even though some state spaces are unreasonable for the current task.
For instance, if two objects are too far from each other for the robot to reach both, it will be reasonable to separate the two objects into different locations instead of merging them into a single one.
However, given the limited planning time, it is almost impossible for S3O-Random to select the most suitable state space especially when there are many objects, thus resulting in much lower task completion rates. 
On the other hand, it is expected to see more Voronoi area merging operations (including feasible and infeasible ones) for S3O-Random than our method which prefers only the feasible ones.
As a result, the S3O-Random agent frequently chooses to navigate only a few times and tries to complete the whole task, which is not ideal.
In comparison, S3O (ours) seeks balance in task completion rate and robot execution time.


\subsection{Real Robot Demonstration}
We demonstrated the generated plan using S3O-GROP$^*$ on a real robot, as shown in Fig.~\ref{fig:real}.
We use the Human Support Robot (HSR) from Toyota~\cite{yamamoto2019development}.
The robot is given a ``tidy home'' task, including collecting three empty cans and moving the apple to the white plate.
Using our planning framework, the robot planned to navigate to the first position to do three manipulation actions: ``collect'' (i.e., pick up the object and put it into a garbage bag mounted on the robot) the green and red cans, and pick up the apple.
While holding the apple in hand, the robot then went to the second position to place the apple on the plate. 
Finally, the robot planned to go to the third position to collect the blue can. 
A demo video has been uploaded as a supplementary file.

\section{Conclusion}
This paper introduces Symbolic State Space Optimization~(S3O), which constructs state space candidates from object-centric partitioning of the configuration space and ranks each candidate by probabilistically evaluating action feasibility values.
S3O is applied to a TAMP system for long horizon mobile manipulation tasks where we further improve motion-level search efficiency using the CMA-ES algorithm.
The resulting framework is called S3O-GROP$^*$, which was extensively evaluated in simulation and demonstrated it using a real robot.
Results showed that S3O-GROP$^*$ produces task-motion plans that are of higher quality than existing TAMP algorithms in terms of task completion rate and robot execution time. 


\section*{Acknowledgements}
This work has taken place in the Autonomous Intelligent Robotics (AIR) group at SUNY Binghamton and the Learning Agents Research Group (LARG) at UT Austin. 
AIR research is supported in part by grants from NSF (IIS-1925044), Ford Motor Company, OPPO, and SUNY Research Foundation.
LARG research is supported in part by NSF (FAIN-2019844), ONR (N00014-18-2243), ARO (W911NF-19-2-0333), DARPA, Bosch, and UT Austin's Good Systems grand challenge.  
Peter Stone serves as the Executive Director of Sony AI America and receives financial compensation for this work.
The terms of this arrangement have been reviewed and approved by the University of Texas at Austin in accordance with its policy on objectivity in research.

\bibliographystyle{IEEEtran} 
\bibliography{ref}

\end{document}